# Honeybees-inspired heuristic algorithms for numerical optimisation

Muharrem Dugenci

Karabuk University, Dept of Industrial Engineering, Karabuk, Turkey

Abstract

Swarm intelligence is all about developing collective behaviours to solve complex, ill-structured and large-scale problems. Efficiency in collective behaviours depends on how to harmonise the individual contributions so that a complementary collective effort can be achieved to offer a useful solution. The main points in organising the harmony remains as managing the diversification and intensification actions appropriately, where the efficiency of collective behaviours depends on blending these two actions appropriately. In this study, two swarm intelligence algorithms inspired of natural honeybee colonies have been overviewed with many respects and two new revisions and a hybrid version have been studied to improve the efficiencies in solving numerical optimisation problems, which are well-known hard benchmarks. Consequently, the revisions and especially the hybrid algorithm proposed have outperformed the two original bee algorithms in solving these very hard numerical optimisation benchmarks.

## 1. Introduction

Collectivism is one of the approaches commonly found useful for problem solving in the modern times. This is motivated by the fact that collective effort pays off better than individual effort in the real life and has been bought in by computer science researchers and implemented in various problem-solving approaches. Swarm intelligence is known to be a family of collectivism-based problem solving frameworks such as ant colony optimisation, particle swarm optimisation, artificial bee colonies etc. imposing use of population of solutions, here-forth called swarm of individuals. The main benefit of population-based metaheuristic approaches, particularly swarm intelligence algorithms, is that the algorithms nicely harmonise local search activities around various neighbourhoods without guaranteeing to cover the whole search space. Therein, the local search is devised, to a certain extend, for intensifying the search and enhancement activities are facilitated to diversify the search for managing change among neighbourhoods.

Diversification plays a crucial role to arrange visiting unseen regions of the search space as efficiently as possible so that the search effort for optimum solution would not be trapped in locality and be able to keep enough energy for further search. On the other hand, intensification is required to make the search algorithm as focus as possible so that any particular local region would remain under-examined. A balanced/well-featured search algorithm harmonises the actions required for both diversification and intensification, which is required for effective and efficient search. In fact, individual solution-driven search algorithms conduct more intensified search while population-driven algorithms are more diversifying by their nature. Hence, swarm intelligence algorithms do require intensification of the search in local regions as they deliver very diverse search by default. This feature applies to the algorithms developed inspiring of honeybees, where a number of bees algorithm (BA) (Pham, et al. 2006) and artificial bee colony (ABC) (Karaboga 2005) variants have been re-designed to manage/handle such a harmony among various search actions. In fact, various hybrid algorithms are devised mainly for this purposes, where a verity of difficult problems can be solved with a more generalised search that well-featured with diverse and focus search activities, adequately (Kong, et al. 2012, Aydin 2012, Yuce, Pham, et al. 2015). However it is observed that the existing mechanics of BA and ABC algorithms do not sufficiently support intensification, which drives us to further investigations.

The main aim of this study is investigate these properties in both BA and ABC and then to seek for an efficient harmony in both algorithms, where both of BA and ABC are revised accordingly and then hybridised into a new algorithm following preliminary investigations. All details are provided in the following sections.

The rest of this paper is organised as follows: Section 2 introduces swarm intelligence algorithms inspired of natural honeybee colonies, while Section 3 examines the two algorithms (BA and ABC) and introduces the suggested revisions for both and proposes a new hybrid algorithm. Section 4 includes a comprehensive experimental study to test the performance of all algorithms, while Section 5 discusses the results in comparison to the existing relevant literature and Section 6 provides the conclusions.

## 2. Swarm Intelligence and Honeybee-inspired Algorithms

Swarm intelligence is one of the cutting-edge soft computing technologies used for solving various optimisation problems in more efficient ways. This is because the approaches and frameworks proposed are adaptive, flexible and robust in the way that the algorithms handle the problems using various techniques of collectivism. Collective effort by each individual within the swarms is managed by sharing the information regarding search activities towards the common targets. That helps divers the search by its nature.

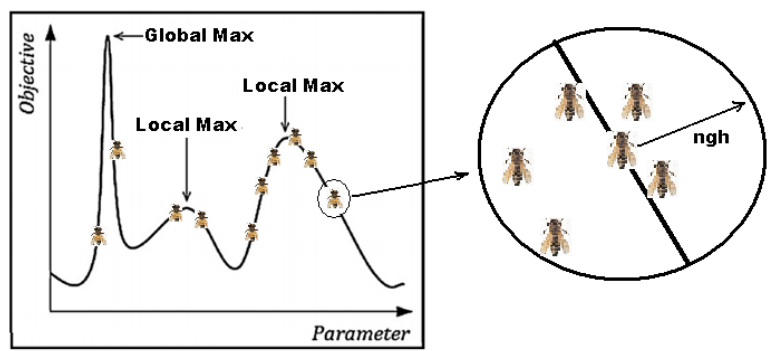

Figure 1; Search metaphorically delivered by honeybees

Figure 1 sketches the search idea metaphorically delivered by honeybees, where a typical function, which has multiple optima points, is explored through for the optimum points, while the search conducted within a neighbourhood by a team of bees is also spotted out to reflect the idea and its implementation. This metaphoric idea has been borrowed from honeybees and their way of collective search by mainly two algorithms, as explained above. Both of the algorithms are detailed in the following subsections.

## 2.1 Bees Algorithm (BA)

Bees algorithm is another mainstream of swarm intelligence algorithms inspired of natural honeybee colonies introduced by Pham and his associates (Pham, et al. 2006, Yuce, Packianather, et al. 2013). It looks like a typical population based optimisation algorithm in which solutions are considered as individual bees and are evaluated based on the fitness-function-like evaluation rules, which are usually of simple objective functions. The algorithm imposes a search procedure inspired of food/nectar exploration process by honeybees within the nature. An elitist approach is followed to search through the most fruitful regions of the search space so that the optimum or a useful near optimum can be found as fast as possible without causing further complexities. This algorithm has not only been used for solving numerical optimisation problems, e.g. benchmark functions, neural network training etc., but also been considered for solving a variety of combinatorial optimisation problems (Keskin, Dugenci and Kacaroglu 2014, Yuce, Pham, et al. 2015).

Let $\mathcal{X}$ be a population of solutions, which is considered to be the bee colony and let $\mathbf{x}_i = \{x_{i,j} | i = 1, \ldots, N; j = 1, \ldots, D\}$ represent solution *i* within this population, which is also called an individual bee as a member of colony/swarm, where *N* denotes the size of bee colony, $N = |\mathcal{X}|$, and *D* is the size of input set. Suppose also that $F(\mathbf{x}_i)$ is a function defined ($f_i : \mathbf{x}_i \to \mathbb{R}$) to measure the quality/fitness of solution $\mathbf{x}_i$. The initial population/swarm of bees is generated using $x_{i,j} = x_{i,min} + \rho * (x_{i,max} - x_{i,min})$, where $x_{i,j}$ is a data point for $j^{th}$ input of $\mathbf{x}_i$ solution initialised to be a random value within the range of $[x_{i,min}, x_{i,max}]$ normalised with the random number of $\rho$.

After generating the initial swarm, each individual bee is evaluated using the fitness function created based on the main objective of the problem tackled. The bees are, then, classified based on their performance/fitness; a set of elite bees, $\mathcal{E}$, where $\mathbf{y}_e \in \mathcal{E}$ and $\mathbf{y}_e = \{y_{e,j} | e = 1, \ldots, |\mathcal{E}|; j \in D\}$, a set of moderate search bees, $\mathcal{M}$, where $\mathbf{z}_m \in \mathcal{M}$ and $\mathbf{z}_m = \{z_{m,j} | m = 1, \ldots, |\mathcal{M}|; j \in D\}$, and a set of employee bees, $\mathcal{J}$, where $\mathbf{x}_k \in \mathcal{J}$ and $\mathbf{x}_k = \{x_{k,j} | k = 1, \ldots, N - (|\mathcal{E}| - |\mathcal{M}|); j \in D\}$. Therefore, $\mathcal{X} = \mathcal{E} \cup \mathcal{M} \cup \mathcal{J}$, where $N_e = |\mathcal{E}|, N_m = |\mathcal{M}|$, and $|\mathcal{J}| = N - (|\mathcal{E}| - |\mathcal{M}|)$. In order for moving to the next generation, $\mathcal{E} \in \mathcal{X}$ and $\mathcal{M} \in \mathcal{X}$ are preserved ahead and the rest of the population, which are employee bees, are scraped.

The next step of producing the next generation is to deploy supporting bees, which are not created initially, but later while breeding the new generation in order for supporting each elite, $\mathbf{y}_e$, and moderate, $\mathbf{z}_m$, bees within the neighbourhood of each. Each individual elite bee, $\mathbf{y}_e$, is supported with a team of bees to further explore within its neighbourhood. This extends the size of elite bees' set from $N_e$ to $N_e \times \beta$ while the moderate search bees are also supported in the same way, but with different predefined supporting team of bees. This also increases the size of moderate bee set to $N_m \times \gamma$, where $\beta$ and $\gamma$ are predetermined fixed numbers, to identify how many bees to be recruited in the neighbourhood of each elite and moderate bee, respectively. The supporting bees, which are deployed in the search regions of elite and moderate bees, are created with the rule of $x_{i,j} = x_{i,j} + \rho * \delta$, where $\rho$ is a random number generated within the range of (-1, 1) and $\delta$ is another predetermined fixed value to be the step-size of change in any input of a solution/ a bee. This rule can be specified for each of the bee types as follows: (i) supporting bees for elite bees with $y_{i,j} = y_{i,j} + \rho * \delta$, while for moderate search bees with $z_{i,j} = z_{i,j} + \rho * \delta$, . Once support teams of bees are deployed within corresponding search regions, the majority of the swarm of the next generation becomes complete. The remaining small portion of the new colony (around 20%) is randomly generated in the way of the initial random population.

Once the elite bees, moderate search bees and the other are identified, the predefined number of supporting bees are sent to each neighbourhood of both types of these bees. This procedure is repeated until a predetermined stopping criterion is met.

## 2.2 ABC algorithm

Artificial Bee Colony (ABC) is another very popular swarm intelligence algorithms developed inspiring of the collective behaviours of honeybee colonies. Karaboga (2005) has first initiated this algorithm to solve numerical optimisation problems (Karaboga and Basturk 2007) and then extended the applications with various combinatorial optimisation ones (Karaboga, Gorkemli, et al. 2014), (Pan, et al. 2011). ABC imposes considering individual solutions as sources of food (nectar) for honey bees and searching around each solution is named to be collective activities of various types of bees. There are mainly two bee types envisaged; Employed and Unemployed, where Unemployed bees can be in two types; Onlooker and Scout bees. A set of search activities is organised around the nectar sources by recruiting various types of bees in various configurations.

Let $\vec{x_i}$ be a solution, defined as an input vector of $\mathcal{D}$ size considered as a source of nectar. A population of $\mathcal{N}$ different sources are initially generated using $x_{ij} = x_{i,min} + \rho * (x_{i,max} - x_{i.min})$, where $i = 1,..,\mathcal{N}; j = 1,..,\mathcal{D}$; $x_{i,min}$ and $x_{i.max}$ are minimum and maximum values of $i^{th}$ input of $\vec{x_i}$ source. Once the whole population of the sources is generated completely, then, the nectar level of each source is determined to identify the quality of each, which becomes the fitness value of each solution. Following this step, the employed bees start operating on each source to search for sources with better quality using $v_{ij} = x_{ij} + \phi_{ij}(x_{ij} - x_{ik})$, where $\vec{v_i}$ is the new source found following an interaction between $\vec{x_i}$ and $\vec{x_k}$, which is a randomly selected known source among many within the colony of the generation. The difference calculated between the two sources is normalised with a randomly generated $\phi_{ij} \in (-a, a)$. After the new source identified, a decision is made whether or not to adopt the new source to replace the original one. The ultimate fitness of a typical source decision is calculated using:

$$F(\vec{x_i}) = \begin{cases} \dfrac{1}{1+f(\vec{x_i})}, & f(\vec{x_i}) \geq 0 \\ 1 + |f(x)|, & otherwise \end{cases}$$

Onlooker bees start operations following complete by Employed bees. The main role of onlooker bees is to monitor the employed bees and taking the search further using a probabilistic process, where a probability of $p_i$ is calculated using $p_i = \frac{F(\vec{x_i})}{\sum_{i=1}^{N} F(\vec{x_i})}$ for each individual candidate source and a roulette-wheel selection rule is used to make a chose a solution for further explorations. The neighbourhood of a chosen source is conducted with $v_{ij} = x_{ij} + \phi_{ij}(x_{ij} - x_{ik})$ similar to employed bees. A small size memory is associated with each further investigated source if any progress is achieved or not. A counter for each investigated source is created and run up to a predefined threshold. If no progress accomplished, then the source is removed from the colony.

Scout bees, then, follow onlookers to diversify the colony, randomly inserting new sources using the initial rule of source generation: $x_{ij} = x_{i,min} + \rho * (x_{i,max} - x_{i,min})$. This generational process is repeated until a certain level of satisfaction is reached. As part of the above-mentioned process, each individual solution/source can be included in the next generation via either of the following cases: (i) a source would remain without any change, (ii) an employed bee would generate a new solution, (iii) an onlooker bee may bring a new solution, (iv) a source would be found by both employed or onlooker bees, or (v) an investigated source is replaced with a new source as a result of non-improvement decision. It is a fact that each solution is attempted for improvement at least once, would be investigated with more attempts if the its fitness remains high.

## 3. Algorithms Revisited

The abovementioned honeybees-inspired algorithms have been examined with respect to the balance between diversification and intensification of the search, and few ideas have been put together for the purpose of improving their performances in solving numerical optimisation problems.

Following structural and experimental analysis, both of the algorithms introduced above have been found with strengths and weaknesses with respect to diversification and intensification of search process. Both ABC and BA algorithms include freshly generated random solutions into the new generations to a certain level, where diversification of the search is achieved in this way. In addition, BA algorithm intensifies the search on fruitful sources, where further search attempts are organised around highly fitted sources/solutions, which helps intensification further, while ABC uses memory-like mechanism to let scout bees intensify their search around certain sources for a number of attempts until it is understood that the source is dried out. Once a source is dried out, it is deleted from the population.

On the other hand, both algorithms conduct search with few shortcomings, which have been considered in this study to enhance their capabilities in these regards. BA algorithm uses a parameter to normalise the step-size, so-called environmental/neighbourhood factor and denoted with $\delta$, above. It is set to a fixed value at the initialisation stage and kept as it is to the end of the search. This makes granularity of the step size coarse-grained in approximating the optimum value, which drifts intensification away, and prevents the search to reach the optimum in most of the time. Another weakness of BA algorithm is the diminishing the probability of having random solutions within the population, especially during the late stages of the search. This causes disability of diversification at later stages. In the case of ABC, the weaknesses arise in two points; (i) the sources taken out of population are evaluated not based on the fitness, but, improvability, which can cause disregard of useful solutions, (ii) in addition to this, some useful and very well-improved solutions can be decommissioned from the population since their improvability is reduced to $0$ according the criteria adopted. Both of these weaknesses can drive the algorithm towards very unfertile region of search space.

A number of ideas have given rise to enhance the capabilities of both of the bee algorithms following the abovementioned structural assessments.

## 3.1 Bees Algorithm Revised (Rev BA)

The main revision envisaged for BA based on the shortcomings discussed above is to make step-sizes more adjustable and fine-tuned. This is identified to be about the fixed-valued (constant) $\delta$ within the update rule, $z_{i,j} = z_{i,j} + \rho * \delta$, where $z_{i,j}$ is a single dimension of a whole solution and $\rho$ is a random number within the range of [-1, 1]. This constant valued parameter, $\delta$, makes the approach coarse-grained, which causes step-size of the change to be not easily adjustable in finer precisions and can take much longer time to approximate. In order to avoid this shortcoming the update rule is revised as follows: $z_{i,j} = z_{i,j} + \rho * \delta * z_{i,j}$, where $\delta$ is made to be a rate within the range of [0,1], and can be adaptive, too. Therefore, the new step-size calculated with $\delta * z_{i,j}$ will be more adjustable and proportional to the range of $(z_{min}, z_{max})$ with which the algorithm can approximate much faster than before, and more preciously. The update rule is applied to all types of bees recruited as part of the algorithm, while the rest of the algorithm remains as original.

## 3.2 ABC Algorithm Revised (Rev ABC)

Following the shortcomings discussed above, two revisions have been envisaged to achieve ABC improvement; one is to collect all results from all employed and onlooker bees and then apply roulette-wheel selection instead of original practice, and the other revision is to adopt a rank-based selection rule for the next generation, where 25% of top ranked solution from entire existing solution set, $\mathcal{N} + \mathcal{E}$, where $\mathcal{N}$ denotes original bee colony and $\mathcal{E}$ is the number of generated solutions.

## 3.3 Hybrid Bees Algorithm (Hybrid)

This hybridisation is managed based on the framework of BA algorithm with implementing not only bee operations from BA algorithm but also all other abovementioned algorithms. This hybrid algorithm systematically harmonises/reuses the equations (1) – (5) for generating new solutions/bees as well as neighbours for the existing elite and fit bees, where equation (1) is used for generating the initial swarm and independently exploring for better nectar sources while equations (2) – (5) are used to send supporting bees around each elite bee.

$$\vec{x_i} = \vec{x}_{min} + \vec{\rho} * (\vec{x}_{max} - \vec{x}_{min}) \quad \text{for } \forall i \in N \tag{1}$$

$$\vec{x_i} = \vec{x_i} + \vec{\rho} * \delta \quad \text{for } \forall i \in N \quad \text{and} \quad \delta \in \mathbb{R} \tag{2}$$

$$\vec{v_i} = \vec{x_i} + \vec{\phi_i}(\vec{x_i} - \vec{x_k}) \quad k \in N \quad \text{and} \quad \text{for } \forall i \in N \tag{3}$$

$$\vec{x_i} = \vec{x_i} + \vec{\rho} * \delta * \vec{x_i} \qquad \text{for } \forall\, i \in N \quad \text{and} \quad \delta \in [0,1] \qquad (4)$$

$$\vec{v_i} = \vec{x_i} + \vec{\phi_i}(\vec{x_i} - \vec{x_k}) \qquad k \in \{Q_1\ of\ N\} \quad \text{and} \quad \text{for } \forall\, i \in N \qquad (5)$$

Equation (2), (3), (4) and (5) are the neighbourhood rules used, respectively, by the ordinary BA algorithm, the revised BA algorithm, ABC and revised ABC algorithms to explore around a local nectar source, which means a local region of the search space in optimisation context. The hybrid algorithm randomly selects one of these rules to generate a neighbouring solution of a particular elite solution, each time, to complete up $\beta$ supporting bees for each elite so that $N_e \times \beta$ bees can be placed in the new generation. The moderate search bees use only equation (2) and (3) for generating their neighbouring solutions to complete $\gamma$ number of supporting bees so as to place $N_m \times \gamma$ solutions in the next swarm while the independent bees explore with equation (1) for further generations of randomly searched nectars. The rest of algorithmic mechanics of ordinary Bee algorithm applies to the hybrid algorithm until a certain satisfactory level is achieved.

# 4. Experimental Results

The following section introduces the experimental study to demonstrate the performance of above-mentioned well-known bee algorithms and the revised ones envisaged to enhance the capabilities via performances. First of all, the performance tests and analysis have been made using the following numerical optimisation benchmarks, which are very well-known benchmarks used by the researchers for these purposes. Obviously, all of these functions are multi-dimensional functions, which can also be considered as many-dimensional functions, where the tests have been conducted over their 5, 30, 60, 100 and 150 dimensionalities. The reason to opt with these dimensions is that the literature (Kong, et al. 2012) (Alam, Islam and Murase 2012, Karaboga and Basturk 2007, Yuce, Packianather, et al. 2013) reports solving these problems with similar dimensions, where 100 and 150 dimensions are exercised first time by this study. Two of the functions are know as uni-model, which means that they have only single optimum points while the other four are multi-model functions meaning that they can have multiple optimum points. These are all well-known and challenging benchmark functions used to test optimisation algorithms across the literature of this field. An extensive study on a number of numerical optimisation benchmarks including those considered below is reported in (Suganthan, et al. 2005).

1. Sphere function
$$f_1(x) = \sum_{i=1}^{D} x_i^2$$

2. Rosenbrock function
$$f_2(x) = \sum_{i=1}^{D} \left[100(x_{i+1} - x_i^2)^2 + (x_i - 1)^2\right]$$

3. Ackley function
$$f_3(x) = -20e^{-0.2\sqrt{\frac{1}{D}\sum_{i=1}^{D} x_i^2}} - e^{(\frac{1}{D}\sum_{i=1}^{D} \cos(2\pi x_i))} + 20 - e$$

4. Griewank function
$$f_4(x) = \frac{1}{4000}\sum_{i=1}^{D} x_i^2 - \prod_{i=1}^{D} \cos\left(\frac{x_i}{\sqrt{i}}\right) + 1$$

5. Rastrigin function

$$f_5(x) = \sum_{i=1}^{D}\left[x_i^2 - 10\cos(2\pi x_i) + 10\right]$$

6. Schwefel function

$$f_6(x) = \sum_{i=1}^{D}(x_i - \sin(\sqrt{|x_i|}))$$

The parametric design details of the algorithms are provided in Table 1, where the parameters of the main three algorithms tabulated. This is due to the fact that the revised versions of both BA and ABC algorithms have the same parametric values since they suggest more procedural rather than parametric changes. As a matter of fact, the neighbourhood structures of the algorithms, which is also a procedural difference, are indicated as follows: all algorithms use fixed-sized local neighbourhood, while BA has a rank-based random selection, ABC uses roulette-wheel selection and, in fact, Hybrid adopts both in a systematic use. In addition, Hybrid algorithm selects mate-bees from top quartile when operating with revised ABC.

Table 1; parametric details of the algorithmic configurations.

|  |  | BA | ABC | Hybrid |
| --- | --- | --- | --- | --- |
| Population/Swarm size | N | 100 | 100 | 100 |
| Number of elite bees | $N_e$ | 5 | -- | 5 |
| Number of moderate search bees | $N_m$ | 20 | -- | 20 |
| Number of bees supporting elite bees | $\beta$ | 40 |  | 40 |
| Number of Independent bees | $|\mathcal{I}|$ | 30 | -- | 30 |
| Neighbourhood factor | $\delta$ | 0.1 | -- | $0.1 * \vec{x_i}$ |
| Non-improvability threshold | L | -- | 200 | -- |

The experimentation has been started with rather lower dimensions and gradually increased up in due course. The first set of experiments has been done with a dimension of 5 for all benchmarks, just to be in-line with the existing literature. The results are tabulated in Table 2, where each benchmark function is introduced with the input range, modality and the optimum value alongside with the performance of each of the 5 algorithms. The same configuration of the table has been repeated for three iteration levels; 200, 1000 and 5000. The performance of the algorithms provided accordingly within each sub-table, where results gained after 200 iteration come at the top section while results gained with 1000 iterations are in the middle section and the 5000-iterations results placed at the bottom section. The same structure applies to Table 3, which presents the results by the same algorithms for 30 dimensions of the functions.

Table 2; Experimental results by all 5 bee algorithms with 3 levels of iterations for 5-D benchmark functions.

|  | D=5 | 200 Iterations |  |  |  |  |  |  |
| --- | --- | --- | --- | --- | --- | --- | --- | --- |
|  | Input Range | Model | Optimum | BA | Rev BA | ABC | Rev ABC | Hybrid |
| **Sphere** | (-100,100) | Uni | 0 | 0.015 | 0 | 2.17E-23 | 4.02E-26 | 0 |

| | | | | | | | |
|---|---|---|---|---|---|---|---|
| Rosenbrock | (-2.048,2.048) | Uni | 0 | 0.022 | 2.209 | 0.93 | 1.061 | 0.432 |
| Ackley | (-32.768,32.768) | Multi | 0 | 0.599 | 4.44E-16 | 1.26E-11 | 5.01E-14 | 4.44E-16 |
| Griewank | (-600,600) | Multi | 0 | 0.156 | 0 | 0.039 | 0.027 | 0 |
| Rastrigin | (-5.12,5.12) | Multi | 0 | 1.085 | 0 | 0.279 | 1.989 | 0 |
| Schwefel | (-500,500) | Multi | -2094.9145 | -1976 | -1752 | -2094 | -1976 | -1858 |
| | D=5 | 1000 iterations | | | | | | |
| Sphere | (-100,100) | Uni | 0 | 0.003 | 0 | 0 | 0 | 0 |
| Rosenbrock | (-2.048,2.048) | Uni | 0 | 0.058 | 0.081 | 0.136 | 0.677 | 2.76E-04 |
| Ackley | (-32.768,32.768) | Multi | 0 | 0.277 | 3.99E-15 | 4.44E-16 | 4.44E-16 | 4.44E-16 |
| Griewank | (-600,600) | Multi | 0 | 0.147 | 0 | 0.027 | 0.027 | 0 |
| Rastrigin | (-5.12,5.12) | Multi | 0 | 1.029 | 0 | 0 | 0.994 | 0 |
| Schwefel | (-500,500) | Multi | -2094.9145 | -2094 | -1968 | -2094 | -2094 | -1976 |
| | D=5 | 5000 iterations | | | | | | |
| Sphere | (-100,100) | Uni | 0 | 2.92E-05 | 0 | 0 | 0 | 0 |
| Rosenbrock | (-2.048,2.048) | Uni | 0 | 0.026 | 3.35E-05 | 3.78E-04 | 0.063 | 0.002 |
| Ackley | (-32.768,32.768) | Multi | 0 | 0.13 | 4.44E-16 | 4.44E-16 | 4.44E-16 | 4.44E-16 |
| Griewank | (-600,600) | Multi | 0 | 0.101 | 0 | 0.019 | 0.019 | 0 |
| Rastrigin | (-5.12,5.12) | Multi | 0 | 1 | 0 | 0 | 1.989 | 0 |
| Schwefel | (-500,500) | Multi | -2094.9145 | -2094 | -2094 | -2094 | -2094 | -2094 |

Table 3; Experimental results by all 5 bee algorithms with 3 levels of iterations for 30-D benchmark functions

| | D=30 | 200 iterations | | | | | | |
|---|---|---|---|---|---|---|---|---|
| | Input Range | Model | Optimum | BA | Rev BA | ABC | Rev ABC | Hybrid |
| Sphere | (-100,100) | Uni | 0 | 170 | **0** | 510 | 4 | **2.41E-25** |
| Rosenbrock | (-2.048,2.048) | Uni | 0 | 38 | 28 | 254 | 28 | **26** |
| Ackley | (-32.768,32.768) | Multi | 0 | 9 | 1 | 7 | 1 | **0.07** |
| Griewank | (-600,600) | Multi | 0 | 2 | **0** | 6 | 1 | **0** |
| Rastrigin | (-5.12,5.12) | Multi | 0 | 112 | **0** | 226 | 44 | **0** |
| Schwefel | (-500,500) | Multi | -12569.487 | -9264 | -9439 | -4678 | **-9536** | -9192 |
| | D=30 | 1000 iterations | | | | | | |
| Sphere | (-100,100) | Uni | 0 | 6 | **0** | 8.00E-05 | 5.87E-16 | **0** |
| Rosenbrock | (-2.048,2.048) | Uni | 0 | 24 | 27 | 26 | **20** | 21 |
| Ackley | (-32.768,32.768) | Multi | 0 | 5 | **4.44E-16** | 0.002 | 1.01E-08 | **4.44E-16** |
| Griewank | (-600,600) | Multi | 0 | 1 | **0** | 0.001 | 2.55E-15 | **0** |
| Rastrigin | (-5.12,5.12) | Multi | 0 | 120 | **0** | 218 | 43 | **0** |
| Schwefel | (-500,500) | Multi | -12569.487 | -9368 | -8723 | -5722 | **-9619** | -8873 |
| | D=30 | 5000 iterations | | | | | | |
| Sphere | (-100,100) | Uni | 0 | 3.64E+00 | **0** | 1.83E-38 | 1.40E-45 | **0** |
| Rosenbrock | (-2.048,2.048) | Uni | 0 | 18 | 27 | 21 | **11** | 17 |
| Ackley | (-32.768,32.768) | Multi | 0 | 4 | **4.44E-16** | 7.50E-15 | 7.54E-15 | **4.44E-16** |
| Griewank | (-600,600) | Multi | 0 | 0.21 | **0** | 0 | 0 | **0** |
| Rastrigin | (-5.12,5.12) | Multi | 0 | 123 | **0** | 195 | 43 | **0** |
| Schwefel | (-500,500) | Multi | -12569.487 | -8836 | -8891 | -5555 | **-10571** | -8876 |

As explained above, the experimental results reported in Table 2 and 3 are the performance of five algorithms for each of the benchmark problems. The dimensions of the problems and the number of iterations gradually growing throughout of these tables as can be seen.

Sphere function is easily solved by almost all algorithms with 5 dimensions over 200 iterations, where both Rev BA and Hybrid solves it with exact optimum, but, BA substantially remains distant, and the others are just approximate with very minor difference. All algorithms approximate to the optimum with both 1000-iteration and 5000-iteration cases. The function with 30 dimensions becomes challenging, where Rev BA finds the optimum, Hybrid hits the optimum with an ignorable difference after 200 iterations, while the other significantly remain distant. After 1000 iterations BA and ABC only stay struggling, but the other three solve the problem with exact solution. BA only remains a little bit distant after 5000 iterations while the rest solve it exactly. Hence, the best achievement so far is by Rev BA, and Hybrid.

Rosenbrock function is one of two function found challenging in this study. None of the algorithms have found the optimum while the best with 5 dimensions is by BA and with 30 dimensions by Rev ABC. Algorithms' performances improve with increasing number of iterations at levels of 1000 and 5000, however, the optimum is still not achieved, although BA and Rev ABC perform better for 5 and 30 dimensions, respectively, and Hybrid always follows as the second best.

The best approximation for Ackley function is made by Hybrid, while Rev BA is competing with Hybrid to reach the exact optimum, however, both remain in a very ignorable distance. In fact, Hybrid performs the best after 200 iterations for 5 dimensions cases, but, Rev BA competes with Hybrid in other cases of 30 dimensions.

Griewank function is best approximated by Rev BA and Hybrid algorithms, even as early as 200 iterations in both dimensions of 5 and 30. The other algorithms solve the problem to the optimum level after 5000 iterations. Rev BA and Hybrid solve Rastrigin function to optimum in both dimensions (5 and 30) after 200 iterations, while the other algorithms struggle to approximate even after 5000 iterations. It is important to indicate that this function is attended by (Kong, et al. 2012) with 5 and 10 dimensions only. Schwefel function remains as the most challenging benchmark since ABC and Rev ABC solve it with 5 dimensions after 200 iterations, but, none of the algorithms managed solving the problems to the optimum with higher dimensions even after 5000 iterations, where initial swarms/populations escalate to very different the results, each time.

Table 4; Experimental results by all 5 bee algorithms with 1000 and 5000 iterations for 60-D benchmarks

|  | D=60 | 1000 Iterations | | | | | | |
|---|---|---|---|---|---|---|---|---|
|  | Input Ranges | Model | Optimum | BA | Rev BA | ABC | Rev ABC | Hybrid |
| **Sphere** | (-100,100) | Uni | 0 | 9.029 | **0** | 1.27E+04 | 3.90E-03 | **2.80E-45** |
| **Rosenbrock** | (-2.048,2.048) | Uni | 0 | 792.867 | 58.882 | 2357.957 | 111.236 | **54.965** |
| **Ackley** | (-32.768,32.768) | Multi | 0 | 19.512 | **3.24E-14** | 15.093 | 2.27E+00 | **3.95E-14** |
| **Griewank** | (-600,600) | Multi | 0 | 855.597 | **0** | 102.018 | 8.48E-03 | **0** |
| **Rastrigin** | (-5.12,5.12) | Multi | 0 | 637.548 | **0** | 616.653 | 200.983 | **0** |
| **Schwefel** | (-500,500) | Multi | -25138.974 | -16169.357 | -8461.699 | 8056.175 | **-17522.662** | -14847.428 |
|  | D=60 | 5000 Iterations | | | | | | |
| **Sphere** | (-100,100) | Uni | 0 | 1.00E+01 | **1.10E-44** | 8.69E+03 | 5.51E-33 | **7.00E-45** |

| Rosenbrock | (-2.048,2.048) | Uni | 0 | 76.018 | 58.722 | 1,963 | 96.026 | **0.0242** |
| Ackley | (-32.768,32.768) | Multi | 0 | 19.227 | 4.44E-16 | 1.38E+01 | 2.78E+00 | **4.00E-16** |
| Griewank | (-600,600) | Multi | 0 | 655.712 | **5.55E-11** | 74.153 | 0.109 | **2.22E-16** |
| Rastrigin | (-5.12,5.12) | Multi | 0 | 402.313 | **0** | 578.02 | 172.127 | **0** |
| Schwefel | (-500,500) | Multi | -25138.974 | -14996 | -12579 | -8554 | **-19540** | -14726 |

Table 4 presents the performances of all 5 algorithms for 60 dimensional benchmarks after 1000 and 5000 iterations, where it is clear that both ordinary BA and ABC algorithms remain very underperforming in comparison to the revised versions and the hybrid algorithm, although their performance improve with more iterations as indicated in the bottom (5000 iteration cases) section of the table. On the other hand, Rev BA, Rev ABC and Hybrid approximate to the optimum in four functions even after 1000 iterations, while struggle in solving Rosenbrok and Schwefel functions, despite that their performance improves in Rosenbrok function. These results indicate that Schwefel function clearly requires far more attention to better approximate. It is also notable that the results after 5000 iterations deviate from the optimum with very ignorable level.

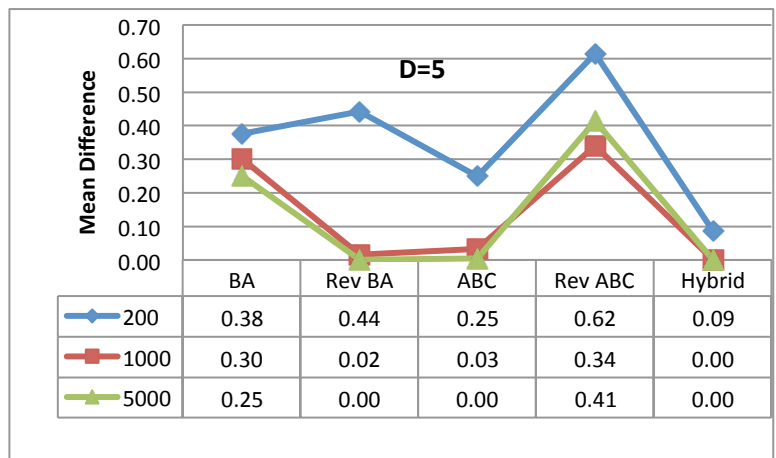

Figure 2(a)

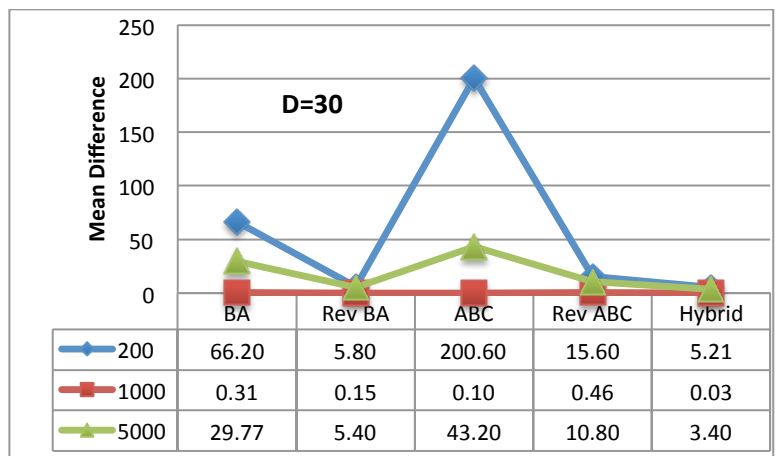

Figure 2(b)

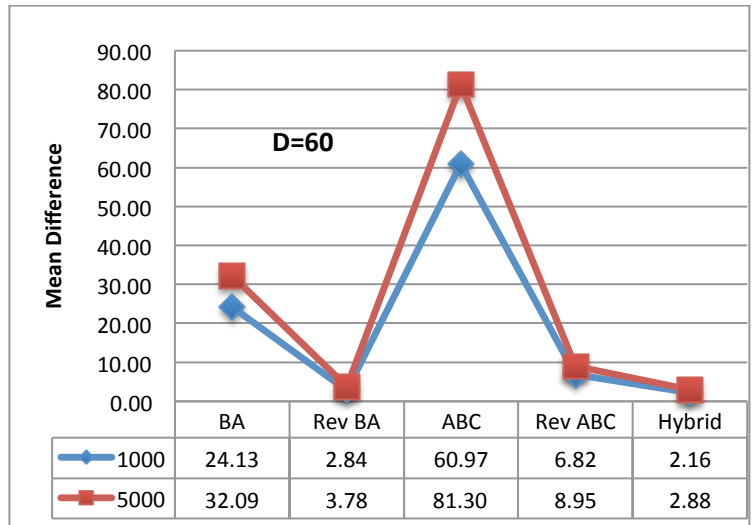

Figure 2(c)

Figure 2 Differences between optimum and results found by the algorithms for (a) D=5, (b) D=30 and (c) D=60.

Figure 2(a), 2(b) and 2(c) present the differences between known optimum values and the achieved results averaged over all benchmark problems categorised dimensions and the number of iterations taken. Figure 2(a) and 2(b) include the results for 200 iterations, but Figure 2(c) does not include since 200 iterations remain too short for growing dimensions. All three figures clearly suggest that Hybrid algorithm outperforms all others and its approximation goes closer to 0. On the other hand, revised algorithms perform better then the original algorithms in the same overall point of view, where Rev BA remains as the first runner algorithm after Hybrid. It is also observed that ABC performs much better when dimension is lower, but, performs not as good as the other rivals with growing dimension. However, Rev ABC, the revised ABC, is one of the competitors with Hybrid regardless of the growing dimensions.

The results in Table 5 and 6 include the performance of BA, ABC and the Hybrid algorithms only, since beyond D=60, the revised algorithms seem to underperform following their original versions. However, Hybrid remains competitive and solving four functions to optimum out of the six benchmarks. Table 5 and 6 presents the results of Hybrid in comparison to original BA and ABC for D=100 and D=150 running the algorithms for 1000 and 5000 iterations. Hybrid solves Sphere, Ackley, Grienwak and Rastrigin functions to optimum after 1000 iterations for all dimensions including 60-D, 100-D and 150-D. However, the algorithms, BA, ABC and the revised versions of these two, remain behind this level of achievement with growing dimensions. The algorithms other than Hybrid seem falling in a local optimum around 20 while solving Ackley function for dimensions of 100 and 150. Rosenbrock function is the second challenging benchmark among all, where the approximation of Hybrid remain just below 100 for 100-D and below 150 for 150-D cases. Clearly, Schwefel function is the most challenging one since the approximation of all algorithms stays far apart of the expected optimum. This hints that Schwefel function requires particular attention. Both BA and ABC improve their performances with increase of iteration numbers from 1000 to 5000, but, the level of improvement, apparently, remains rather weak. That means that the approximation of both algorithms approach to the ultimate level of achievement, and beyond this level of iterations a significant improvement is not expected.

Table 5; Experimental results for 100-D cases with iterations of 1000 and 5000.

| | D=100 | 1000 iterations | | | | |
|---|---|---|---|---|---|---|
| | Input Range | Model | Optimum | BA | ABC | Hybrid |
| **Sphere** | (-100,100) | Uni | 0 | 169.88 | 95,853.60 | **0.00** |
| **Rosenbrock** | (-2.048,2.048) | Uni | 0 | 219.58 | 14,169.54 | **96.80** |
| **Ackley** | (-32.768,32.768) | Multi | 0 | 20.53 | 19.45 | **3.99E-15** |
| **Griewank** | (-600,600) | Multi | 0 | 1242.29 | 909.12 | **0.00** |
| **Rastrigin** | (-5.12,5.12) | Multi | 0 | 808.55 | 1,314.08 | **0.00** |
| **Schwefel** | (-500,500) | Multi | -41,898.29 | -26,548.66 | -9,065.48 | **-26,671** |
| | D=100 | 5000 iterations | | | | |
| **Sphere** | (-100,100) | Uni | 0 | 61.08 | 72,499.42 | **0.00** |
| **Rosenbrock** | (-2.048,2.048) | Uni | 0 | 160.66 | 10,679.20 | **92.52** |
| **Ackley** | (-32.768,32.768) | Multi | 0 | 19.49 | 19.05 | **3.99E-15** |
| **Griewank** | (-600,600) | Multi | 0 | 0.90 | 768.32 | **0.00** |
| **Rastrigin** | (-5.12,5.12) | Multi | 0 | 764.34 | 1,265.31 | **0.00** |
| **Schwefel** | (-500,500) | Multi | -41,898.29 | -26,057.94 | -9,255.25 | **-26,537.00** |
| | D=150 | 10000 iterations | | | | |
| **Sphere** | (-100,100) | Uni | 0 | 2.52E-07 | 72,499.42 | **0** |
| **Rosenbrock** | (-2.048,2.048) | Uni | 0 | 92.08 | 10,554.82 | **86.78** |
| **Ackley** | (-32.768,32.768) | Multi | 0 | 18.98 | 19.02 | **6.79E-16** |
| **Griewank** | (-600,600) | Multi | 0 | 2.03E-07 | 610.62 | **0** |
| **Rastrigin** | (-5.12,5.12) | Multi | 0 | 564.14 | 1,210.09 | **0** |
| **Schwefel** | (-500,500) | Multi | -62,847.435 | -26,065.43 | -10,746.90 | **-26,750.00** |

Table 6; Experimental results for 150-D cases with iterations of 1000 and 5000.

| | D=150 | 1000 iterations | | | | |
|---|---|---|---|---|---|---|
| | Input Range | Model | Optimum | BA | ABC | Hybrid |
| **Sphere** | (-100,100) | Uni | 0 | 28289.01 | 200,742.50 | **0.00** |
| **Rosenbrock** | (-2.048,2.048) | Uni | 0 | 623.26 | 30,157.64 | **144.27** |
| **Ackley** | (-32.768,32.768) | Multi | 0 | 20.51 | 20.50 | **3.99E-15** |
| **Griewank** | (-600,600) | Multi | 0 | 2232.72 | 2,186.48 | **0.00** |
| **Rastrigin** | (-5.12,5.12) | Multi | 0 | 1421.92 | 22,142.83 | **0.00** |
| **Schwefel** | (-500,500) | Multi | -62,847.435 | -38,277.94 | -10,974.98 | **-37,851,645** |
| | D=150 | 5000 iterations | | | | |
| **Sphere** | (-100,100) | Uni | 0 | 0.02 | 221,512.73 | **1.10E-44** |
| **Rosenbrock** | (-2.048,2.048) | Uni | 0 | 142.89 | 32,620.37 | **140.68** |
| **Ackley** | (-32.768,32.768) | Multi | 0 | 19.18 | 20.47 | **1.47E-15** |
| **Griewank** | (-600,600) | Multi | 0 | 2212.12 | 1,995.48 | **0.00** |
| **Rastrigin** | (-5.12,5.12) | Multi | 0 | 868.31 | 2,106.19 | **0.00** |
| **Schwefel** | (-500,500) | Multi | -62,847.435 | -35,174.20 | -12,060.97 | **-38,568.57** |
| | D=150 | 10000 iterations | | | | |
| **Sphere** | (-100,100) | Uni | 0 | 2.50E-06 | 200,945.00 | **0** |
| Rosenbrock | (-2.048,2.048) | Uni | 0 | 139.42 | 24,671.02 | **137.33** |
| Ackley | (-32.768,32.768) | Multi | 0 | 19.11 | 20.24 | **1.46E-15** |
| Griewank | (-600,600) | Multi | 0 | 1.96E-06 | 1,631.69 | **0** |
| Rastrigin | (-5.12,5.12) | Multi | 0 | 829.79 | 2,071.09 | **0** |
| Schwefel | (-500,500) | Multi | -62,847.435 | -38,506.52 | -12,092.69 | **-38,568.57** |

Both Table 5 and 6 include the results gained after 10000 iterations for both large dimensional cases, i.e. 100-D and/or 150-D. The results do not include any surprise on that beyond a certain number of iterations; the achievement is not improving significantly, where Hybrid evidently out performs both of its competitors. In fact both of BA and ABC algorithms approximate very roughly, while Hybrid approaches to the optimum values except the cases of Rosenbrok and Schwefel functions.

Figure 3(a), 3(b), 3(c) and 3(d) present the overall performance indication of BA, ABC and Hybrid algorithms for the dimensions of 100 and 150. Similar to the previous cases depicted in Figure 2, the results of all the algorithms for all functions have been further processed to calculate the differences between the optima and the results produced, and then averaged accordingly. Figure 3(a) and 3(b) plots the averaged differences for 100-D, while Figure 3(c) and 3(d) plots for 150-D. It is observed that, as suggested by Figure 3(a) and 3(c), ABC significantly underperforms in comparison to both BA and Hybrid, while Figure 3(b) and 3(d) compare the performance of BA with Hybrid, where Hybrid significantly outperforms BA. The performance of BA improves with increasing number of iterations while Hybrid looks approximated to a steady state as suggested by Figure 3(b) and 3(d) noting that the averaged difference by Hybrid looks more substantial in 100-D cases than 150-D ones.

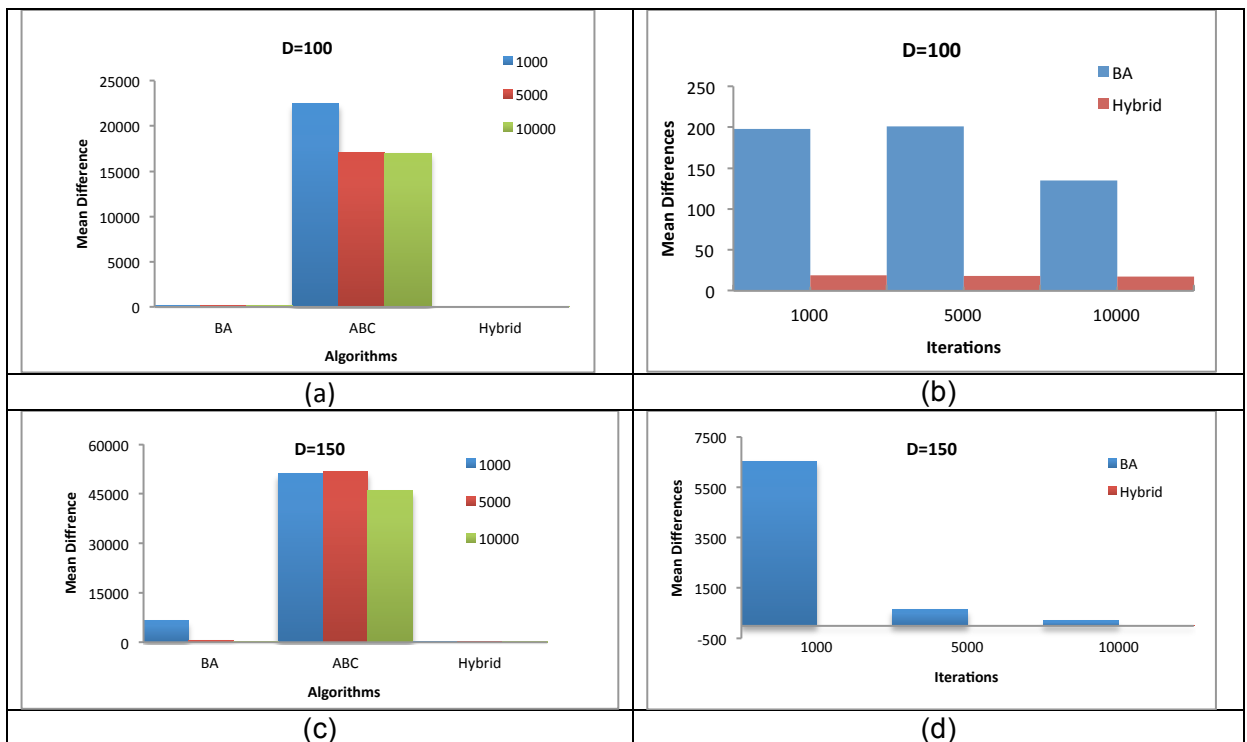

Figure 3; Averaged overall achievements by BA, ABC and Hybrid; (a) comparative results for 100-D cases (b) comparative results by BA and Hybrid only, for 100-D cases, (c) comparative results by all three for 150-D cases, (d) comparative results by BA and Hybrid only, for 150-D cases.

## 5. Relevant Works

A variety of works have considered the same benchmark functions as ours for testing the performance of their versions of BA and ABC algorithms. Kong, et al. (2012) report their study in which this work has inspired of to initiate. It imposes an orthogonal initialisation for their hybrid algorithm, where it uses 60 dimensional problems at most, and mostly underperformed in comparison to our results. Hacibeyoglu, Kocer and Arslan (2012) produced results for the same set of benchmarks, but did not tabulate them so as to be compared with our results, where the level of dimensions is clearly lower than our case. Kiran and Gunduz (2012) have borrowed and embedded a crossover operator from genetic algorithms to solve the numerical benchmark functions, where they considered 50 dimensions at most and the results seem to be very fluctuating as standard deviations are higher then mean statistics in some cases. Karaboga and Basturk (2007) have first published their ABC algorithm with the same set of benchmark numerical functions solving them in relatively lower dimensions. However, Karaboga and Akay (2009) have presented the success of the same algorithm providing extensive details of their comparative study, where they solved around 50 benchmarks including our benchmarking problems. The algorithms seems implemented very successfully for dimensions up to 30 noting that many other ABC implementations could not hit that level of success. Kiran and Findik (2015) present a directed/adaptive ABC algorithm solving the benchmarks with 10, 30 and 50 dimensions, where our results are competitive with them at this level while we solve the problems for much higher dimensions.

On the other hand, Pham et al. (2006) introduces their BAs algorithm with solving the same set of benchmarking numerical function with rather very lower dimensions, e.g. up to D=10. Likewise, Yuce et al. (2013) have also attempted to solve a number of benchmark functions including those considered in this study with up to 10 dimensions at most. Hussein, Sahran and Abdullah (2014) have improved BAs algorithm with a pre-processing of particular initialisation algorithm and gained better results than both of (Pham, et al. 2006) and (Yuce, Packianather, et al. 2013) in solving the same set of benchmarks with up to 60 dimensions, where our results apparently outperform for all functions except Schwefel.

A number of other metaheuristic and/or swarm intelligence algorithms have also attempted to solve the benchmarks we considered, recently. Based on the relevance that the same functions have been attempted, it is decided to include these studies in the review to help grasp the difficulty of the problems attended. Gong, et al. (2011), Liu, Niu and Luo (2015), Zhao and Tang (2008) and Xin, et al. (2010) have published their results for the benchmark problems up to 30 dimensions using different variants of particle swarm optimisation, differential evolution and a particular algorithm so called monkey algorithm. Their results are apparently either not better than, or remain competitive with ours. Likewise, Han et al. (2014), Rehmani and Yusof (2014) and Alam et al. (2012) have introduced their approaches for 30 and 50 dimensions, where our approach usually outperforms them or remain competitive. None of the following references have attempted dimensions larger than 50, but, the majority of them have only considered up to 30, while our approach outperform them in major (Guo, et al. 2014, Piotrowski 2015, Kashan 2015, Dogan and Olmez 2015, Alam, Islam and Yao 2011). These studies have mostly compared their result with those produced by Suganthan, et al. (2005) in which a comprehensive study is extensively reported on solving a number of numerical optimisation benchmarks.

# 6. Conclusions

In this study, a comprehensive investigation is conducted to review the capabilities of Ba and ABC algorithms with respect to diversification and intensification in their search conduct. Both frameworks have been comparatively tested in solving very high-dimensional numerical optimisation benchmarks. Revisions have been proposed for each algorithm for performance enhancement purposes. The results clearly suggested that revised versions of both BA and ABC (Rev BA and Rev ABC) outperformed the original algorithms by large. Furthermore, a hybrid algorithm based on the original and their revised versions has been developed and tested, accordingly, resulting that the hybrid algorithm significantly improves the performance in solving very high-dimensional numerical optimisation benchmarks. This achievement is attained with better harmony induced in the hybrid algorithm, where both of Rev BA and Rev ABC provided better intensification and randomly and systematically use of operators helped achieve improved diversification. This hybrid version is going to be tested for combinatorial optimisation problems as the next step of this research.